\documentclass[10pt,sigconf,screen,letterpaper]{acmart}
\AtBeginDocument{%
  }
    
\setcopyright{none}
\acmConference[arXiv'23]{}{October 9, 2023}{Online}

\acmPrice{}
\acmDOI{}
\acmISBN{}
\acmYear{}

\makeatletter
\renewcommand{\@copyrightowner}{Copyright held by the owner/author(s). 
This is the author's version of the work.
It is posted here for your personal use. Not for redistribution. 
}
\makeatother

\newcommand{\para }[1]{\vspace{2mm} \noindent  {\bf #1}}
\newcommand{\1}{{\em (i)}}
\newcommand{\2}{{\em (ii)}}
\newcommand{\3}{{\em (iii)}}
\newcommand{\4}{{\em (iv)}}

\newcommand{\cmark}{\cellcolor{green!25}$\checkmark$}
\newcommand{\xmark}{\cellcolor{red!25}$\times$}

\usepackage{xspace}
\usepackage{multirow}
\usepackage{graphicx}
\usepackage{subcaption}
\usepackage{enumitem}
\usepackage{titlesec}
\usepackage{pifont}
\usepackage{xcolor}
\usepackage{colortbl}
\usepackage{hyperref}

\newcommand*\ttvar[1]{\texttt{\expandafter\dottvar\detokenize{#1}\relax}}
\newcommand*\dottvar[1]{\ifx\relax#1\else
  \expandafter\ifx\string_#1\string_\allowbreak\else#1\fi
  \expandafter\dottvar\fi}

\setlist[itemize]{noitemsep,topsep=0pt,leftmargin=.12in, after=\vspace{2pt}} 

\titlespacing*{\subsection}{0pt}{1pt}{0.2\baselineskip}
\titlespacing*{\section}{0pt}{5pt}{0.2\baselineskip}
\titlespacing*{\subsubsection}{0pt}{5pt}{0.2\baselineskip}

\begin{document}
\sloppy

\title{Mobile AR Depth Estimation: Challenges \& Prospects Extended Version} 

\author{Ashkan Ganj}
\orcid{0009-0006-3490-0471}
\email{aganj@wpi.edu}
\affiliation{%
  \institution{Worcester Polytechnic Institute}
}

\author{Yiqin Zhao}
\orcid{0000-0003-1044-4732}
\email{yzhao11@wpi.edu}
\affiliation{%
  \institution{Worcester Polytechnic Institute}
}
\author{Hang Su}
\orcid{}
\email{
hangsu@nvidia.com}
\affiliation{%
  \institution{Nvidia Research}
  \department{}
}

\author{Tian Guo}
\orcid{0000-0003-0060-2266}
\email{tian@wpi.edu}
\affiliation{%
  \institution{Worcester Polytechnic Institute}
}

\renewcommand{\shortauthors}{Ganj et al.}

\begin{abstract}
Metric depth estimation plays an important role in mobile augmented reality (AR). With accurate metric depth, we can achieve more realistic user interactions such as object placement and occlusion detection. 
While specialized hardware like LiDAR demonstrates its promise, its restricted availability, i.e., only on selected high-end mobile devices, and performance limitations such as range and sensitivity to the environment, make it less ideal.
Monocular depth estimation, on the other hand, relies solely on mobile cameras, which are ubiquitous, making it a promising alternative for mobile AR. 

In this paper, we investigate the challenges and opportunities of achieving accurate metric depth estimation in mobile AR. 
We tested four different state-of-the-art monocular depth estimation models on a newly introduced dataset (ARKitScenes) and identified three types of challenges: hardware, data, and model related challenges. Furthermore, our research provides promising future directions to explore and solve those challenges. These directions include \1 using more hardware-related information from the mobile device's camera and other available sensors, \2 capturing high-quality data to reflect real-world AR scenarios, and \3 designing a model architecture to utilize the new information.
\end{abstract}
\maketitle

\section{Introduction}
Depth estimation is pivotal to mobile AR as it enables more realistic interaction between virtual and real-world objects. Relative depth estimation provides relative distances of pixels with respect to the camera, while \emph{metric depth estimation} estimates these distances in metric units like meters.
Recent breakthroughs in computer vision have yielded impressive outcomes in relative depth estimation by relying on large-scale data training and specialized scale-invariant loss functions~\cite{birkl2023midas, wu2022toward}.

However, relative depth often falls short of addressing mobile AR's unique requirements, e.g., the appropriate scales for object placement.
Consequently, accurate \emph{metric depth} is needed. However, despite producing reasonable metric depth estimation on specific datasets, existing works often can't generalize well to other datasets and hence real-world scenarios~\cite{yin2023metric, tri-zerodepth}.

Monocular and single-image metric depth estimation solutions are popular options for mobile AR due to low hardware requirements and ease of use. In this paper, through empirical evaluations, we identify and categorize challenges and opportunities to achieve better metric depth for mobile AR.
Our analysis of the ARKitScenes~\cite{dehghan2021arkitscenes} reveals interesting attributes of mobile AR scenarios such as diverse object types, LiDAR sensor limitations, unpredictable user movements, and fragmented views. Our evaluations of several SOTA metric depth estimation methods~\cite{https://doi.org/10.48550/arxiv.2302.12288, tri-zerodepth,wu2022toward} on the ARKitScene have revealed obvious performance gaps on this real-world mobile dataset, although these models have reported impressive generalization performance on tested datasets.

For example, our analysis shows that ZoeDepth~\cite{https://doi.org/10.48550/arxiv.2302.12288}, a model we evaluated in depth, struggles to perform well on real-world mobile AR data and exhibits problems like catastrophic forgetting on the NYU Depth V2 dataset ~\cite{Silberman:ECCV12} after training on ARKitScenes.
Furthermore, our analysis shows that these models face significant generalization issues due to reasons like diverse camera parameters, making adaptation to different scenes at deployment challenging.

Our analysis pinpoints important and unique challenges faced by metric depth estimation in mobile AR. 
First, mobile AR devices often receive insufficient metric information. Most devices lack specialized hardware like LiDAR or ToF sensors for direct metric measurements. Even among those with LiDAR, such as certain high-end iPhones, the provided metric data is sparse, complicating the creation of dense depth maps. 
Second, the unpredictability of mobile AR user movements introduces significant dynamics to camera data. These user behaviors can produce challenging viewpoints, frames with fragmented views, and limited overlap. Depth estimation models, which rely heavily on contextual cues to determine metric scales, struggle to interpret such irregular data.
Finally, our evaluation indicates that adapting learned metric scale information to real-world mobile AR camera data can be unreliable due to scale ambiguity and difficulties in generalizing across different cameras.

We propose three promising directions to improve metric depth estimation for mobile AR. 
First, we propose two directions to leverage the new opportunities presented by the recent advances in mobile hardware.
Second, we identify four essential characteristics for capturing mobile AR-specific datasets.

Finally, we sketch some model architecture changes needed to utilize the new information. 
To achieve accurate metric depth estimation for mobile AR, one needs to consider all three aspects, hardware, data, and model, collectively.
In summary, our main contributions are as follows: 

\begin{itemize}
    \item Through comprehensive analysis and evaluation on ARKitScene, we reveal unique challenges that metric depth estimation faces in real-world mobile AR applications.
    \item  We conduct an evaluation of a set of SOTA learning-based estimation models. Our results show that even with recent large deep models like ZoeDepth~\cite{https://doi.org/10.48550/arxiv.2302.12288}, achieving accurate metric depth estimation still remains challenging.
    \item  We identify a set of promising directions for the research community to develop future metric depth estimation solutions via insights developed from the current landscape of depth estimation and our analysis.
\end{itemize}

\section{Mobile AR Depth Estimation}
\label{background_mobileAR}

\begin{figure}[t]
    \centering
        \includegraphics[width=\linewidth]{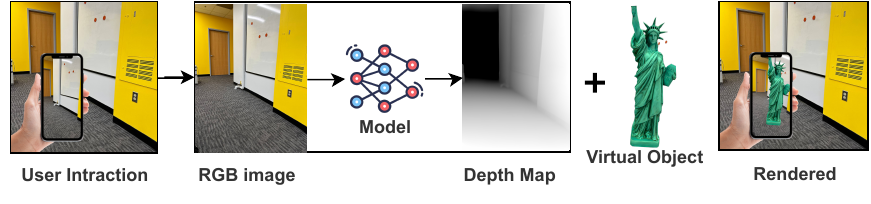}
    \caption{
    An example use case of depth estimation.
    }
\label{fig:mobileAR_workFlow}
\vspace{-0.5cm}
\end{figure}

Depth estimation is a fundamental task in computer vision where the primary goal is to determine the distance of each pixel in an image from the camera lens. Depth estimation includes metric depth and relative depth. \emph{Metric depth} estimation seeks to find the accurate measurement, in an absolute physical unit, of depth values between camera and objects. The resulting depth map reflects actual physical distances, usually in metric units. \emph{Relative depth} estimation determines the order of pixels in a scene without quantifying the exact distances in metric units. It helps understand which objects are closer or further away relative to each other but doesn't provide precise metric depth values.

Mobile AR is one of the major use cases of depth values~\cite{10.1145/3517260}. Figure~ \ref{fig:mobileAR_workFlow} shows a workflow of depth in AR task. Both types of depths are useful but metric depth has more real-world applications than relative depth. Metric depth helps the user to have more accurate and realistic interaction with the world, e.g., via virtual object placement. 
Metric depth information is vital for placing an object in 3D space such that its orientation, scale and perspective are properly adjusted.

Current strategies to solve metric depth problem include:
\1 \emph{Stereo Vision:} Utilizing dual cameras on smartphones to compute depth through disparity~\cite{10.1145/3495243.3560517}. 
\2 \emph{Sensor Based}: Directly measuring depth through light reflections (LiDAR, ToF), available on some high-end devices~\cite{AppleARkit,10.1145/3517260}.
\3 \emph{Single and Monocular Depth:} Employing trained neural networks to predict depth from single or a sequence of images~\cite{https://doi.org/10.48550/arxiv.2302.12288, sayed2022simplerecon}.
The first two approaches often rely on specialized hardware or high-complexity operations to extract information from the additional inputs.

In this paper, we focus on single and monocular depth due to its flexibility and reliance only on ubiquitous cameras, and other reasons detailed in \S\ref{subsec:lidar_limitaitons} and \S\ref{sensor_availability}. 
The inherent difficulty in metric depth estimation arises from the transformation of a 3D scene into a 2D projection captured by a camera. This transformation leads to various challenges that we describe in \S\ref{sec:challenge}. If we had such metric information, depth estimation task would be easier. However, because of how cameras' image formation process work, these information loss is inevitable.

\subsection{The Current Landscape}

\begin{table*}[!t]
\centering
\caption{Comparisons of recent depth estimation models.
None of these works address all three challenges and often do not provide generalization.
}
\label{tab:paperSurvey}
\resizebox{0.8\textwidth}{!}{
\begin{tabular}{@{}l|cc|ccc|c@{}}
\toprule
 & \textbf{Input Type} & \textbf{Output} & \multicolumn{3}{c|}{\textbf{Challenges}} & \textbf{ZeroShot}\\ 
 & & & \textbf{Data-Related} & \textbf{Model-Related} & \textbf{Hardware-Related} & \\
\midrule
\textbf{MiDaS~\cite{birkl2023midas,Ranftl2020}}                     & Single Image & Relative & \cmark & \cmark & \cmark & \cmark \\ 
\textbf{DPT~\cite{Ranftl2021}}                                      & Single Image & Relative & \xmark & \cmark & \xmark & \cmark \\ 
\midrule
\textbf{ZeroDepth~\cite{tri-zerodepth}}                             & Monocular    & Metric   & \xmark & \xmark & \cmark & \cmark \\ 
\textbf{DistDepth~\cite{wu2022toward}}                              & Single Image & Metric   & \cmark & \xmark & \cmark & \cmark \\ 
\textbf{Metric 3D~\cite{yin2023metric}}                           & Monocular    & Metric   & \xmark & \xmark & \cmark & \cmark \\ 
\textbf{AdaBin~\cite{9578024}}                 & Single Image & Metric   & \xmark & \xmark & \cmark & \xmark \\ 
\textbf{Local Bin~\cite{10.1007/978-3-031-19769-7_28}}              & Single Image & Metric   & \xmark & \cmark & \cmark & \cmark \\ 
\textbf{InDepth~\cite{10.1145/3517260}}                             & Sensor Based & Metric   & \cmark & \xmark & \xmark & \xmark \\ 
\textbf{MobiDepth~\cite{10.1145/3495243.3560517}}                   & Stereo       & Metric   & \cmark & - & \cmark & \xmark\\ 
\bottomrule
\end{tabular}
}
\end{table*}

We systematically survey the recent(last 5 years) depth estimation work, as summarized in Table~\ref{tab:paperSurvey}, from the computer vision community, and those specifically targeted at mobile AR.
We find that learning-based depth estimation models, without considering the mobile AR characteristics, often fall short in addressing the challenges (\S\ref{sec:challenge}), even works that specifically created for mobile depth estimation~\cite{10.1145/3517260,10.1145/3495243.3560517} couldn't address all the challenges.

\para{Relative depth estimation} offers the advantage of dealing with pixel distances in a relational manner, without the need for metric units. This concept makes it easier to train models on large and diverse datasets, therefore improving robustness and generalizability. For instance, MiDaS~\cite{birkl2023midas,Ranftl2020} and DPT~\cite{Ranftl2021} have made progress in tackling hardware(\S\ref{subsec:hardware_challenges}) and model-related (\S\ref{subsec:model_challenges}) challenges. They introduced a novel scale-invariant loss function, enabling training on datasets captured with a variety of hardware devices. Nevertheless, these approaches fall short in satisfying the need of AR applications. 

\para{Metric depth estimation} aims to provide exact depth values in real-world units (e.g., meters). This is critical in applications where precise object localization is necessary. Works like ZoeDepth~\cite{https://doi.org/10.48550/arxiv.2302.12288}, AdaBin~\cite{9578024}, LocalBin~\cite{10.1007/978-3-031-19769-7_28} tried to solve metric depth problem by only focusing on single input RGB image and they used methods like formulating the metric regression task as classification or utilizing the genearlizability of relative depth on metric depth. On the other side works like Metric3D~\cite{yin2023metric} and ZeroDepth~\cite{tri-zerodepth} tried to use camera specific information both during inference and training to make the metric depth more accurate.

Our evaluations(\S\ref{sec:limitations_STOA_models}) showed that most of these works don't have a good performance on the ARKitScene dataset.

\para{Mobile-specific depth estimation}  Recent interests from academia and industry. A few recent works including InDepth~\cite{10.1145/3517260}, and MobiDepth~\cite{10.1145/3495243.3560517} attempted to make the models perform well mobile devices by exploring the ToF sensors, reducing the complexity of using dual cameras, and providing a software library for real-time depth. While these works are a step in the right direction for supporting mobile AR, they still do not fully address all the unique challenges. In this paper, we identify promising directions that address all three challenges simultaneously.
\
\subsection{Limitations of Models}
\label{sec:limitations_STOA_models}

\begin{table}[b]
\vspace{-13pt}
\caption{
Comparison of different models' metric depth performance on ARKitScenes.
}
\label{tab:metrics}
\resizebox{\columnwidth}{!}{
\begin{tabular}{@{}l|rrr@{}}
\toprule
\textbf{Model} & \textbf{RMSE}	$\downarrow$ & \textbf{AbsRel}	$\downarrow$  & \textbf{\# of Parameters $\downarrow$}\\ \midrule
    ARKit Depth Completion~\cite{AppleARkit} & 0.04 & 0.02 &Unknown\\
    ZoeDepth~\cite{https://doi.org/10.48550/arxiv.2302.12288} (pre-trained)  & 0.61 & 0.33 &344.82M\\
    ZeroDepth~\cite{tri-zerodepth} (pre-trained) & 0.62 & 0.37 & 233M \\
    DistDepth~\cite{wu2022toward} (pre-trained) & 0.94 & 0.45 & 69M \\
    \midrule
    ZoeDepth (train with MiDaS) & 0.26 & 0.17 & 344.82M \\
    ZoeDepth (with frozen MiDaS) & 0.37 & 0.25 & 344.82M  \\

    \bottomrule
    \end{tabular}
}

\end{table}

In this section we summarize our findings about the performance of SOTA models on the ARKitScenes~\cite{dehghan2021arkitscenes}---a newly introduced dataset designed to highlight the challenges in mobile AR scenarios. 
Table~\ref{tab:metrics} shows the comparisons. 
We selected three models for our study: ZoeDepth~\cite{https://doi.org/10.48550/arxiv.2302.12288}, DistDepth~\cite{wu2022toward}, and ZeroDepth~\cite{tri-zerodepth}. 
For all models, we evaluated their pre-trained versions on a subset of ARKitScenes dataset;The ARKitScenes subset consists of 39K training data and 5.6K validation data, following the original training/validation division, and ground truth depths.
We used the two common metrics \emph{RMSE}, root mean squared error, and \emph{AbsRel}, absolute relative error, for evaluating depth estimation. Despite their claims of being tested as zero-shot cross validation or generalization, we find that these models fall short compared to ARKit performance on a new dataset like ARKitScenes.

As Table~\ref{tab:metrics} shows, ARKit's proprietary depth completion model achieves the best performance with an RMSE of  \(0.04\) and an AbsREL of \(0.02\). A significant factor behind ARKit's superiority is its reliance on specialized hardware-derived (LiDAR) priors. 

We discuss the limitation of LiDAR-based approaches in \S\ref{subsec:lidar_limitaitons}. Table~\ref{tab:metrics} highlights the limitations of current SOTA models in replicating ARKit depth estimation accuracy. Models such as ZoeDepth, ZeroDepth and DistDept have significantly higher RMSE and AbsREL values than ARKit, making them less suitable for mobile AR. Diving further into the details, the various ZoeDepth versions show disparities in performance. We tried two configurations of the ZoeDepth and trained them separately on ARKitScene dataset and even after training the results are worse than ARKit. Also, we observe that ZoeDepth \emph{trained with MiDaS} is more capable in learning features from the dataset. This variance among the ZoeDepth versions underscores the challenge of developing a universally effective model for depth estimation in mobile AR environments. Another key insight obtained from the table is the non-linear relationship between model complexity and performance outcomes. Despite its high parameter count, ZoeDepth's performance isn't in line with its complexity. In comparison, ZeroDepth, with a significantly smaller model size, performs in the same performance range as the more complex ZoeDepth variant.

Our findings challenge the assumption that models trained with a large number of datasets or more complex models result in better depths. We believe more comprehensive approach is necessary for accurate metric depth. The approach should focus on all aspects of depth estimation, such as the quality of the input data, metric clues and camera information, as well as designing more mobile-specific models. The details of promising directions will be outlined in \S\ref{sec:opportunities}.

We summarized some of the reasons for this substantial accuracy gap exhibited by SOTA depth estimation models. More details are in \S\ref{sec:challenge}.

\begin{itemize}
    \item \textbf{Data Specialization}: There lacks high-quality mobile AR datasets for learning-based methods. SOTA models aren't fine-tuned with mobile AR datasets, leading to their inferiority to ARKit.
    \item \textbf{Camera Overfitting}: Models that are trained in certain camera configurations fail to generalize well in scenarios with other camera configurations.
    \item \textbf{Model Architecture}: For single-image metric depth estimation, it is not clear that more complex architectures translate to higher estimation accuracy. 

    \item \textbf{Scene Coverage:} The real-world estimation might use images of partial scene as input, as showed by the ARKitScene. This scene coverage characteristic is in contrast to traditional Computer Vision datasets models are trained on.

\end{itemize}

Overall the results of Table~\ref{tab:metrics} underscore the need for designing a better end-to-end pipeline, which could be achieved by considering all the pipeline aspects like input level change, model level changes, and capturing more specific datasets for mobile AR.

\section{Mobile AR Metric Depth Challenges}
\label{sec:challenge}

We identify three broad challenges via our evaluations in the context of mobile AR. 
The hardware-related challenges (\S\ref{subsec:hardware_challenges}) describe the difficulty stems from and caused by mobile hardware.
\S\ref{subsec:data_challenges} details the data-related challenges in terms of availability and unique characteristics of mobile AR. 
Finally, in \S\ref{subsec:model_challenges}, we pinpoint the SOTA models' limitations. 

\subsection{Hardware-Related Challenges}
\label{subsec:hardware_challenges}

Existing depth estimation models often rely on visual sensory information, e.g., captured by RGB cameras, to obtain depth information. This section describes the challenges pertain to sensor availability, Diverse Camera Models, and the limitations of specialized hardware.

\subsubsection{Sensor Availability.}
\label{sensor_availability}
Mobile hardwares are inherently heterogeneous and not all mobile devices come with the same set of sensors. 
Currently, specialized depth-sensing technologies like LiDAR and ToF sensors are not uniformly available across smartphones and wearable devices~\cite{10.1145/3495243.3560517}. For instance, LiDAR technology is commonly found in Apple's latest products, while ToF sensors are relatively uncommon on the current market. This technological imbalance results in a large subset of devices lacking access to these advanced sensors. 
Consequently, to support depth estimation for a wide range of mobile devices, it is critical to only rely on sensors that are ubiquitous. However, ubiquitous sensors often lack the ability to \emph{directly} measure depth information, as opposed to specialized sensors like LiDAR and ToF. 
This necessitates us to derive all pixel-wise depth information entirely from the visual cues, such as in monocular depth estimation models.  
These models have the advantage of utilizing a single camera or other device-native information for effective operation. As such, monocular depth estimation is a more cost-efficient and universally applicable alternative, fulfilling diverse requirements of mobile AR scenarios.

\subsubsection{Diverse Camera Parameters.}
\label{camera_overfiting}

Mobile AR often requires working with devices with different camera parameters. 
Various camera parameters introduce two major problems for monocular and single depth estimation: 
\begin{itemize}
    \item 
    \textbf{Scale Ambiguity:} This is largely a byproduct of varying camera parameters and models. For instance, when two cameras capture the same object positioned at an identical distance, the resulting scales in the captured images can differ due to the cameras' unique specifications. An object might appear larger in an image captured by a camera with a shorter focal length compared to one with a longer focal length. This disparity leads to the scale ambiguity problem that complicates learning-based depth estimation. 

    \item \textbf{Model Overfitting to a Single Camera Model:}
    Depth estimation models tend to become highly optimized for specific camera characteristics that the training dataset uses. 
    Several works, such as CamConv~\cite{Facil_2019_CVPR}, Metric3D~\cite{yin2023metric}, and ZeroDepth~\cite{tri-zerodepth}, have investigated this problem and tried to solve the camera overfitting problem (more details in\S\ref{sec:using_hardware}).  A common observation from these studies is the models' tendency to overfit to the specific camera model of their training data. This leads to poor generalization to new scenarios. Primary reason for this is that most models are trained on relatively limited datasets---capturing a wide variety of scenes with different cameras is logistically challenging and time-consuming. This makes these models less adaptable to \emph{in-the-wild} situations featuring diverse camera parameters. Therefore, some works attempted to utilize camera parameters. Among the various intrinsic camera parameters, focal length has emerged as a particularly critical factor for accurate depth estimation. 
\end{itemize}

To conclude, despite efforts to train models on more extensive and diverse datasets to tackle generalization, limited generalizability remains an issue. 
Consequently, these models still require architectural adjustments to become adaptable across different camera configurations, restricting their practical utility in real-world applications.

\begin{figure}[t]
    \centering
        \includegraphics[width=\linewidth]{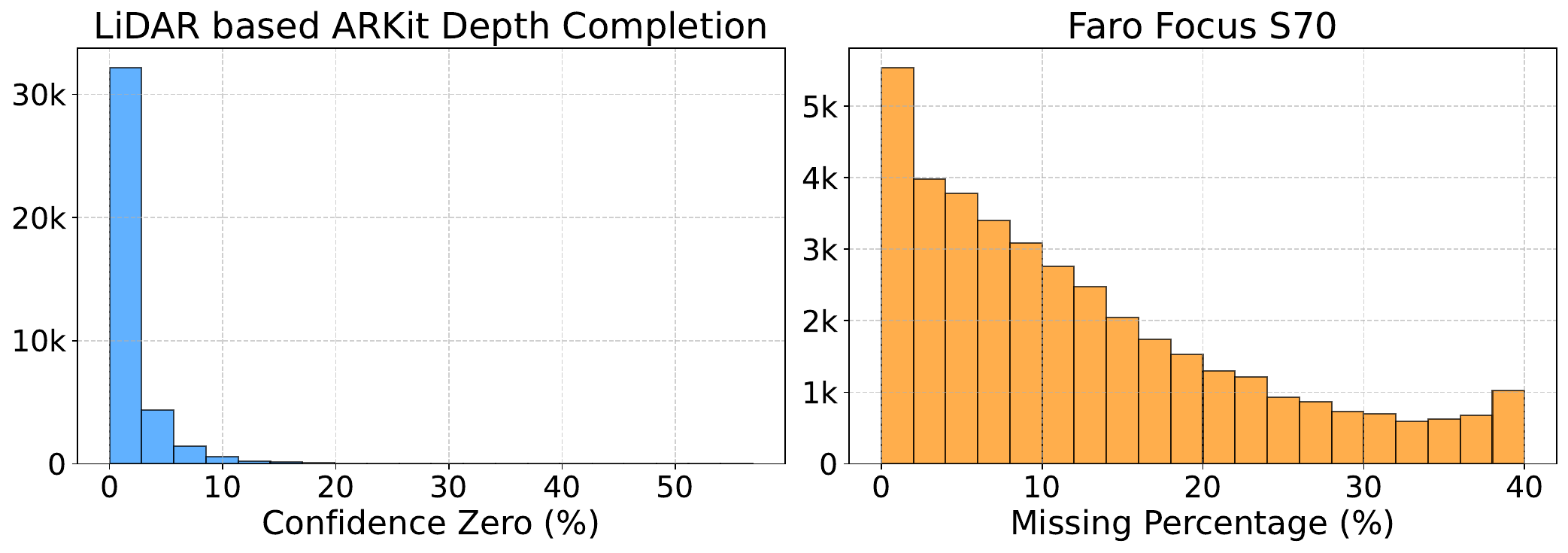}
    \caption{
     Missing depth analysis on ARKitScenes. 
    }
    \label{fig:arkitscenes_stats}
    \vspace{-5mm}
\end{figure}

\begin{table*}[t]
\centering
\caption{Comparison between current indoor datasets for depth estimation.}

\label{tab:indoorDatasets}
\resizebox{1.8\columnwidth}{!}{
\begin{tabular}{@{}l|rrrrr@{}}
\toprule
\textbf{Dataset}    & \textbf{Type}  & \textbf{\# Scenes} & \textbf{\# Images} & \textbf{Captured with mobile} & \textbf{Annotation} \\ 
\midrule
\textbf{NYU Depth V2}~\cite{Silberman:ECCV12} & Indoor         & 464                & 407K               & No                            & RGB-D \\ 
\textbf{DLML Indoor}~\cite{cho2021dimlcvl} & Indoor         & -                  & 220K               & No                            & RGB-D \\ 
\textbf{TUM RGBD}~\cite{sturm12iros}    & Indoor         & -                  & 80K                & No                            & RGB-D \\ 
\textbf{ScanNet}~\cite{dai2017scannet}     & Indoor         & 707                & 2.5M               & No                            & RGB-D \\ 
\midrule
\textbf{ARkitScenes}~\cite{dehghan2021arkitscenes} & Indoor         & 1661               & 450K               & Yes                           & RGB-D \\ 
\textbf{PhoneDepth}~\cite{9857018}  & Indoor/outdoor & 4833 Indoor/1202 Outdoor & 6K & Yes & RGB-D/Stereo \\
\bottomrule
\end{tabular}
}
\end{table*}

\subsubsection{Specialized sensor's Limitations.}
\label{subsec:lidar_limitaitons}
Specialized senors have gained popularity for some of the high-end mobile devices like LiDAR in Apple Devices. In this section we list identified problems mostly by focusing on LiDAR. Methods that rely on LiDAR, such as Apple's ARKit, are constrained by the need for specialized LiDAR hardware, which is not universally present across all mobile devices. In addition to hardware dependency, LiDAR sensors come with their own set of inherent limitations, as discussed bellow:

\begin{itemize}
    \item \textbf{Limitations in Range:} 
    LiDAR sensors have an inherent limitation in their operational range, which is further influenced by environmental and lighting conditions. 
    For example, the operational range of an iPhone's LiDAR is up to 5 meters.
    More specialized depth cameras, such as the Intel RealSense D455f~\cite{intelRealScenes}, extend this range only slightly to 6 meters. These operational ranges fall short when considering datasets like ARkitScenes, including environments with dimensions up to \emph{42 meters}. The limited sensor range thus poses a significant challenge in capturing comprehensive scene details, particularly in larger spaces.
    
    \item \textbf{Sensitivity to Surface Materials and Lighting Conditions:}
    LiDAR and ToF~\cite{10.1145/3517260} sensors are sensitive to the material properties of the objects they measure and the lighting conditions of the environment~\cite{intelRealScenes}. Reflective or dark surfaces can absorb or scatter the light beams, leading to inaccurate depth measurements. Similarly, varying room lighting conditions can affect the sensor's effective range and accuracy.
    
    \item \textbf{Variability in LiDAR Sensor Patterns:}
    LiDAR sensors exhibit variability in their emitted light patterns~\cite{9635718}. This variability can result in significant performance degradation when a depth completion model trained on one LiDAR sensor is applied to another. Consequently, adapting a depth completion model to a different LiDAR sensor would require complete retraining of the model, a process that is both time-consuming and resource-intensive.

\end{itemize}

Our experimental findings on the ARKitScenes dataset support these limitations. Both the specialized ground truth laser scanner and the ARKit depth completion algorithm exhibited significant variability in missing data across different testing scenarios (see Figure \ref{fig:arkitscenes_stats}). The ground truth depth sensor data shows a wide range in the missing data percentage, from 0\% to approximately 40\%, with an average missing percentage of about 12.55\%. On the other hand, the ARKit confidence map has has a lower number of low confidence points, with an average of approximately 1.63\%. However, both sensors show a wide spread of missing data percentages, indicating unreliability. The causes of this missing data could be due to a variety of factors, as discussed above. 

\subsection{Data-Related Challenges}

The selection of appropriate datasets is crucial in training and validating depth estimation models.

An ideal dataset for mobile AR research should capture the unique challenges and reflect the unique characteristics of mobile AR. This section describes challenges associated with data availability (\S\ref{subsub:data_challenge_avail}) and different inference cases (\S\ref{subsub:data_challenge_real_world_AR}). We discuss how to address these challenges by more meticulously designing and capturing the mobile AR datasets in \S\ref{capturingData}.

\subsubsection{Data Availability.} 
\label{subsub:data_challenge_avail}

Our analysis of existing depth datasets (Table~\ref{tab:indoorDatasets}) reveals that there are not a lot of datasets specifically captured for mobile AR. Therefore, researchers will not be able to evaluate their work effectively, and models will not be able to learn the special challenges and hard cases of mobile AR, preventing good generalization.  We summarized a few popular indoor datasets in Table~\ref{tab:indoorDatasets}. Although various datasets exist for indoor depth estimation, most fail to address the specific challenges posed by mobile AR scenarios. For example, datasets such as NYU Depth V2, DLML Indoor, TUM RGB-D, and ScanNet are predominantly captured using specialized or static camera setups, thereby lacking the variability in camera parameters and environmental conditions typically encountered in mobile AR. 

Among the available options, only PhoneDepth and ARKitScenes are captured using mobile devices. PhoneDepth includes both indoor and outdoor scenes.

ARKitScenes~\cite{dehghan2021arkitscenes} is uniquely created for indoor environments and is captured exclusively using mobile devices (an iPad Pro). ARKitScenes includes 1,661 distinct scenes and over 5,048 RGB-D sequences. It also provides high-quality ground truth data sourced from the Faro Focus S70. It is the first RGB-D dataset captured using Apple's phone-based LiDAR sensor, incorporating both ARKit-estimated depth values and high-resolution ground truth depth maps. Below we summarize ARKitScenes's unique characteristics:

\begin{itemize}
    \item \textbf{Practicality for Mobile AR Scenarios:} The dataset's capture methodology, using an iPad, leads to a unique collection of data that closely mimics real-world user interactions with mobile AR applications.

    \item \textbf{Diversity of Indoor Scenes:} ARKitScenes encompasses a wide array of indoor settings, featuring diverse objects, materials, lighting conditions, and spatial configurations. This makes it particularly valuable for the development of robust depth estimation models for real-world mobile AR.
    
    \item \textbf{Comprehensive Depth Information:} The dataset includes ARKit-generated depth information alongside high-resolution ground truth data, facilitating evaluation of depth estimation accuracy.

\end{itemize}

\subsubsection{Difficult Inference Cases.}
\label{subsec:data_challenges}
\begin{figure}[t]
    \centering
    \begin{subfigure}{0.2\textwidth}
        \centering
        \includegraphics[width=0.9\linewidth]{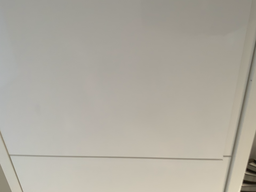}
        \caption{Partial Scene Coverage}
    \end{subfigure}
    \begin{subfigure}{0.2\textwidth}
        \centering
        \includegraphics[width=0.9\linewidth]{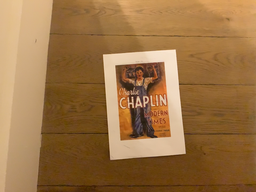}
        \caption{Nested Image}
    \end{subfigure}
    \begin{subfigure}{0.2\textwidth}
        \centering
        \includegraphics[width=0.9\linewidth]{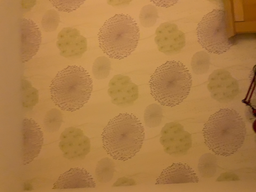}
        \caption{Challenging Object}
    \end{subfigure}
    \begin{subfigure}{0.2\textwidth}
        \centering
        \includegraphics[width=0.9\linewidth]{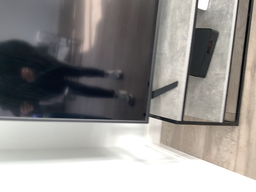}
        \caption{Reflective Object}
    \end{subfigure}
    \caption{Example difficult cases for metric depth estimation from ARKitScenes Dataset.}
    \label{fig:arkitscenes_difficult}
    \vspace{-0.5cm}
\end{figure}

\label{subsub:data_challenge_real_world_AR}
Depth estimation in mobile AR presents specific challenges that makes it more difficult than general use cases. The ARKitScenes dataset provides an excellent representation of these unique challenges. We summarize the distinctive challenges that we found in the ARKitScenes dataset, illustrated in Figure~\ref{fig:arkitscenes_difficult}.

\begin{itemize}
    \item \textbf{Partial Scene Coverage:} Unlike other datasets such as KITTI outdoor dataset~\cite{Geiger2013IJRR} or NYU Depth V2~\cite{Silberman:ECCV12}, where most of the images contain a large portion of the scene, ARKitScenes frames may only contain a portion of the scenes. For example, some ARKitScenes images might show only a portion of a wall or section of the floor. This partial scene coverage reflects the reality of mobile AR scenarios, where a full view of the environment is not always available. 
    The low scene observation could limit the performance of depth estimation models, which often rely heavily on contextual information of input images. 

    \item \textbf{Nested Images:} Another unique aspect of the ARKitScenes dataset is that it contains images with nested objects, such as paintings or pictures within the main scene. This presents a difficulty for single image depth estimation models, as they might attempt to estimate the depth for the nested objects as well.

    \item \textbf{Challenging Object Types:} ARKitScenes contains a wide variety of objects, including challenging elements such as mirrors, white cabinets, and wallpapers with intricate patterns. These objects naturally make the depth estimation task more complex, as they introduce unique visual cues and occlusion patterns that can be difficult for the models to interpret correctly.
\end{itemize}

In summary, compared to other datasets, ARKitScenes possesses data that better reflects the unique challenges associated with depth estimation in mobile AR.
Further, the combination of partial scene coverage, nested images, and the presence of diverse and challenging object types in the ARKitScenes dataset makes it a good candidate to represent the complexity of indoor AR.
Therefore, we select this dataset for our experiments and evaluations to analyze the performance of current SOTA models.

\subsection{Model-Related Challenges}
\label{subsec:model_challenges}

\begin{table}[t]
\caption{ZoeDepth's performance on NYU Depth V2 after training on ARKitScenes.}
\label{tab:zoe_trained_nyu}
\resizebox{0.9\columnwidth}{!}{
\begin{tabular}{@{}l|rrr@{}}
\toprule
\textbf{ZoeDepth Version} & \textbf{RMSE}	$\downarrow$ & \textbf{AbsRel}	$\downarrow$ \\ \midrule
ZoeD-M12-N (pre-trained) & 0.27    & 0.075 \\
ZoeD-ARKitScenes (train with MiDaS) & 0.61  & 0.15  \\
ZoeD-ARKitScenes (with frozen MiDaS) & 0.64 & 0.17  \\
\bottomrule
\end{tabular}
}
\vspace{-0.4cm}
\end{table}

We perform an in-depth analysis of the newly introduced ZoeDepth~\cite{https://doi.org/10.48550/arxiv.2302.12288} to demonstrate the main problems of existing models. 

ZoeDepth builds on top of the widely used relative depth model ,MiDaS~\cite{birkl2023midas,Ranftl2020} by introducing metric heads to attempt to address the generalization issue for metric depth. 
While ZoeDepth's performance has been validated on widely-used datasets like NYU-V2 for indoor scenes and KITTI for outdoor scenarios with the protocol of \emph{zero-shot dataset testing}, we find that ZoeDepth falls short in adapting to the mobile AR scenarios.  
All experiments were conducted with one Nvidia A100; models were trained with a batch size of 8 and for 5 epochs.

\subsubsection{Performance Gap.}
As shown in \S~\ref{sec:limitations_STOA_models}, ZoeDepth and other STOA depth estimation models struggle to achieve comparable metric depth accuracy to ARKit on ARKitScenes. 

Despite ZoeDepth's generalization claim, we see that ARKit consistently outperforms ZoeDepth in both metrics across all training configurations we tested. 
The best-performing ZoeDepth version (trained with MiDaS) has the smallest performance gap to ARKit, with RMSE difference of 0.22 and AbsRel difference of 0.15. 
However, even this ZoeDepth version can suffer from the catastrophic forgetting problem, as discussed in  \S\ref{CatastrophicForgetting}.

\subsubsection{Metric Heads Problem in Learning.}
\label{zoeDepthEval}

\begin{figure}[t]
    \centering
    \begin{subfigure}{0.47\textwidth}
        \includegraphics[width=\linewidth]{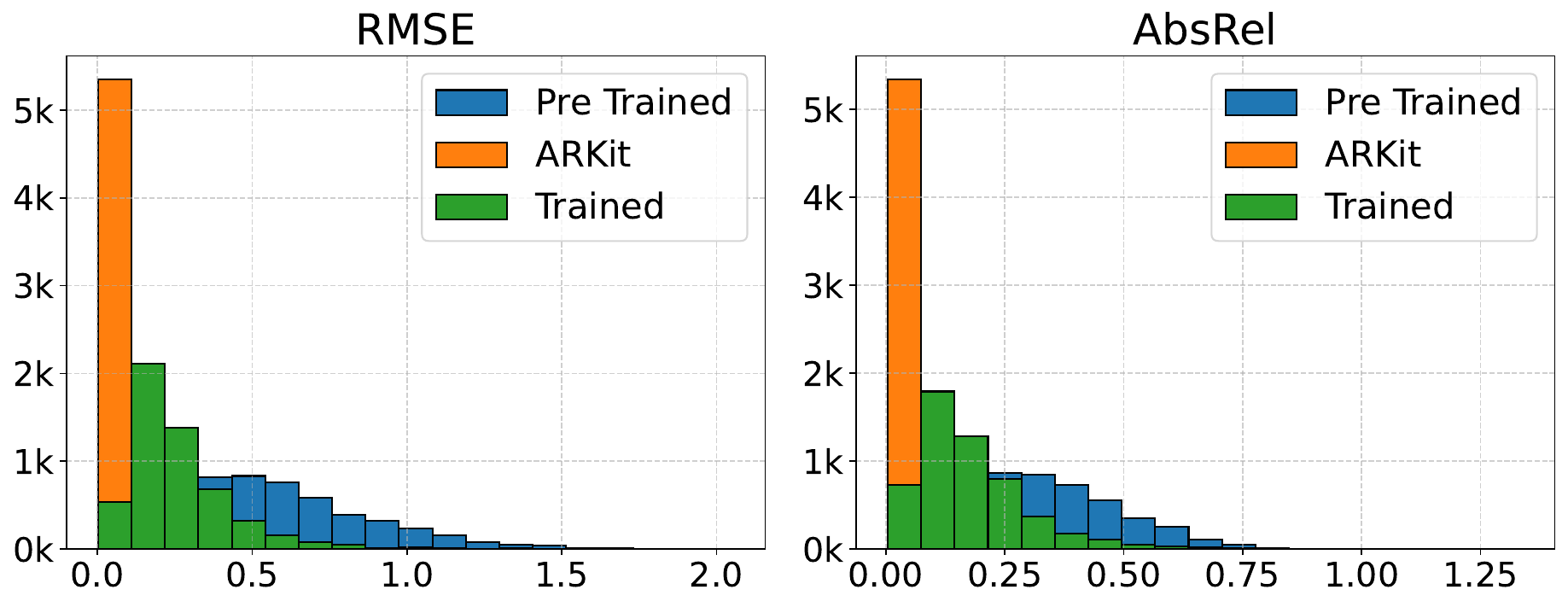}
        \caption{Train together with MiDaS}
        \label{fig:Stats_midas_train}
    \end{subfigure}
    \qquad
    \begin{subfigure}{0.47\textwidth}
        \includegraphics[width=\linewidth]{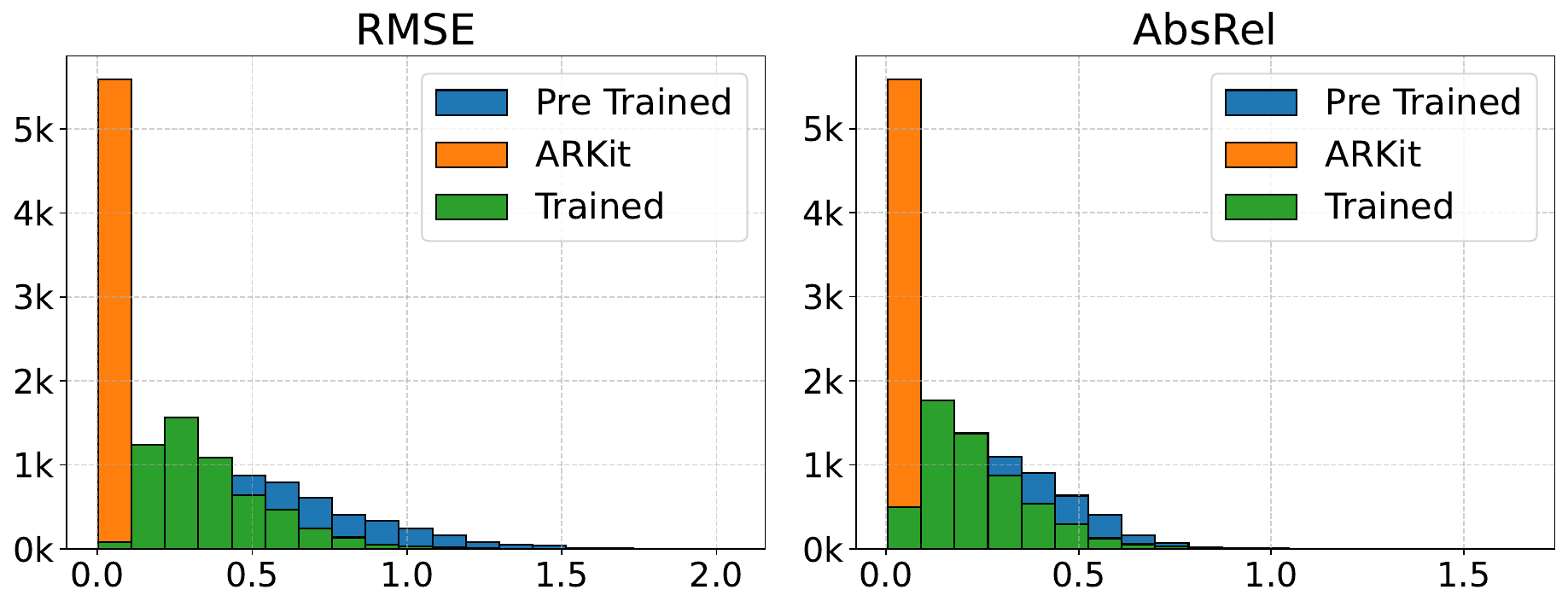}
        \caption{Train with frozen MiDaS}
        \label{fig:Stats_midas_freeze}
    \end{subfigure}
    \caption{Histogram comparisons of ZoeDepth models.
    }
    \label{fig:Comparison_midas}
    \vspace{-6mm}
\end{figure}

We trained two versions of ZoeDepth on the ARKitScenes.
In the first version, we unfreeze all the weights associated with the MiDaS component and jointly train it with the Metric Head. 
For the second version, we freeze the pre-trained MiDaS weights and only train the Metric Heads. 
Figure~\ref{fig:Comparison_midas} depicts the comparisons. 
Our evaluations reveal a significant difference in performance between these two versions. Specifically, training with MiDaS achieves a mean RMSE of \(0.262\) and a mean REL of \(0.179\), significantly outperforming the version trained with frozen MiDaS, with a mean RMSE of \(0.368\) and a mean REL of \(0.244\). 

These results show that the effectiveness of ZoeDepth is heavily influenced by the MiDaS module, highlighting the limitations of the Metric Heads design in capturing metric information from only RGB images.

\subsubsection{Catastrophic Forgetting In Cross-Domain Training.}
\label{CatastrophicForgetting}
To assess the impact of domain-specific training, we re-evaluated ZoeDepth's performance on the NYU Depth V2 dataset using models trained on the ARKitScenes dataset. The results are summarized in Table~\ref{tab:zoe_trained_nyu}. Interestingly, the retrained models fail to show any improvement. Specifically, the RMSE and REL metrics worsen from \(0.27\) and \(0.075\) to approximately \(0.61\) and \(0.15\), respectively. This reveals an important limitation: the model appears to forget its previous training when trying to adapt to a new dataset such as ARKitScenes, thus undermining its generalizability across different domains. Our observation suggests that we can't solve the generalization problem \emph{only with more datasets}.

\section{Opportunities for Better Metric Depth}
\label{sec:opportunities}

\subsection{Utilizing Mobile Hardware}
\label{sec:using_hardware}
Depth estimation based solely on RGB data presents substantial challenges, particularly given the uncertainties and diversities inherent in mobile AR.
In this section, we present two directions by leveraging recent hardware available on mobile devices to improve metric depth accuracy.

\subsubsection{Utilizing Camera Parameters.}
\label{utilize_cam_paramters}
Monocular depth estimation models are often highly dependent on the camera used to capture their training data. As a result, achieving a generalized performance across multiple environments and hardware configurations can be very challenging. Many current methods aim to mitigate this issue by either using large number of training dataset with special loss functions like ZoeDepth~\cite{https://doi.org/10.48550/arxiv.2302.12288} or by integrating camera parameters in some capacity during model training like ZeroDepth~\cite{tri-zerodepth}. While these approaches provide some improvements, they still present limitations for real-world deployment. 
We suspect that the image formation process can contribute to the \emph{scale ambiguity problem}, which makes models sensitive to camera parameters~\cite{yin2023metric}.
One way to mitigate this problem is to provide more information about the image formation process to the models.
Usage of camera parameters can be useful in both input~\cite{yin2023metric, brazil2023omni3d} or intermediate-level~\cite{tri-zerodepth} to the model. 

For example, Metric3D~\cite{yin2023metric} and Omni3D~\cite{brazil2023omni3d} lean towards the use of a \emph{new camera space} as a common framework that allows for more consistent depth estimation, irrespective of the camera used for data collection. The researchers achieved two goals by transforming all training data into this uniform coordinate system: \1 minimizing the model's dependence on the unique intrinsics of individual cameras; \2 they were able to train the model using a large number of datasets without having to worry about scale ambiguity to achieve generalization.

We plan to improve monocular depth estimation by incorporating camera parameters like intrinsic and extrinsic values. These parameters will be collected during inference and used to preprocess camera frames into a new virtual camera space. Processed images will feed into a camera-parameters aware DL model, enhancing its accuracy. Additionally, this strategy will enable more effective data augmentations during training, increasing the model's robustness

\subsubsection{Expanding Observations.}

\begin{table}[t]

\caption{Results of ZoeDepth pre-trained model on different views on NYU Depth V2. }
\label{tab:zoe_crop}
\begin{tabular}{@{}l|rrrr@{}}
\toprule
\textbf{Crop Percentage} & \textbf{0\%} & \textbf{25\%} & \textbf{50\%} & \textbf{75\%} \\ \midrule
RMSE $\downarrow$ & 0.27         & 0.42        & 0.87        & 1.48        \\
AbsRel $\downarrow$ & 0.075        & 0.12        & 0.27        & 0.42        \\
\bottomrule
\end{tabular}
\vspace{-0.3cm}
\end{table}

Our analysis reveals that the quality of depth estimation is significantly impacted by the field of view. Reducing scene coverage, from 100\% to 25\%, can notably limit the effectiveness of SOTA depth estimation models, as observed in our preliminary experiment on the NYU Depth V2 dataset (see Table \ref{tab:zoe_crop}). 

This observation motivates us to explore modern mobile devices features to increase depth accuracy by expanding observations.

\begin{itemize}
    \item \textbf{Utilizing Consecutive Mobile Frames:} In mobile AR applications, scene understanding plays a major role. 
    Because most monocular depth estimation models rely on information gathered from the environment and scene, we believe that improving this understanding can lead to better accuracy. 
    One way to increase scene and environment understanding is to leverage the device's capabilities and user frequent movement in AR session to capture multiple consecutive frames. 

    These frames can then be stitched together to improve the scene's coverage. Our hypothesis is that this approach will offer a more comprehensive depth understanding, eliminating the limitations associated with a restricted field of view. This strategy serves a purpose similar to that of ultra-wide cameras but is particularly useful for devices that lack such hardware features. 
    \item \textbf{Ultra-Wide Camera Integration:} 

    Another way is to leverage the increasingly common ultra-wide cameras in modern mobile devices to expand observations. 
    These ultra-wide cameras offer an expanded field of view, providing the model with additional crucial depth cues like object size and edges. By incorporating this broader scene coverage into the model, we hypothesize that the depth estimation algorithms will be better equipped to identify informative features essential for accurate depth prediction. By exploring both these approaches---consecutive frame stitching and ultra-wide camera utilization---we aim to address the inherent limitations in scene coverage that impact depth estimation. 

\end{itemize}

\subsection{Designing Model Architecture}

In order to more effectively utilize the additional information discussed in the previous section, we will have to design more specialized models for mobile AR. 
Since our goal is to achieve a realistic, real-time AR experience, the new model should also be lightweight. Below we sketch out one potential design as an example is based on the key concept of \emph{depth from focus/defocus}.

Depth that captured by depth from focus/defocus models are usually more reliable than pure deep learning because they can calculate points' depth based on the focus value using Gaussian lens equation~\cite{fujimura2022deep}. The results are based on the specific camera model and image formation process and are often considered as accurate metric depth. 
One challenge this method faces is for texture-less objects and surfaces.
Researchers usually try to fix this problem by using DL models to fill in the missing depth values in the depth map~\cite{Defocus_2020_CVPR, Wang-ICCV-2021}.
However, these methods need to rely on a specific camera, which is problematic (As discussed in \S\ref{camera_overfiting}).
A promising solution, as demonstrated in ~\cite{fujimura2022deep}, is solving this problem by using a new data input for the model which considers the relationship between scene depth, defocus images, and camera settings. We also want to incorporate the method that we mentioned at \S~\ref{utilize_cam_paramters}, which We believe providing more reliable metric information to the model the model can be very useful.

\subsection{Capturing Real-World Datasets}
\label{capturingData}

Existing datasets, as we demonstrated earlier, do not adequately represent the challenges inherent to mobile AR. We believe that creating datasets captured specifically with mobile devices can allow researchers to use more relevant data to enhance the performance of cutting-edge models, thus bridging the gap between model performance and real-world deployment. 

Some essential characteristics an effective mobile AR-specific dataset should possess are: 
\1 \emph{Captured with Mobile Devices} to sure authenticity and real-world applicability.
\2 \emph{Diverse Indoor Scenes Coverage} to represent various scenarios and conditions.
\3 \emph{Accompanied by Additional Information} including camera matrix, device position, etc., for use as priors to improve model estimation accuracy.
\4 \emph{Covering a scene with different FoVs} to enable comprehensive testing and robustness across different scene views.

Some initiative efforts~\cite{10.1145/3615452.3617941} aim to improve the process of capturing mobile scenario-specific data, which can speedup the development of AR-specific datasets.

\section{Conclusion}
In this paper, we performed a systematic study on the current single and monocular depth estimation models in the context of mobile AR. 
Our goal is to understand the current performance gaps between common benchmarks and the real-world scenarios. 
Despite their claim of good generalizability, all tested SOTA models fall short on the mobile-specific ARKitScenes dataset.
Through literature survey and empirical evaluation, we revealed current challenges and problems in mobile AR depth estimation.
To bridge the performance gap, we identified three promising directions that call for better utilization of new mobile hardware features, designing a model architecture to take advantage of the additional input, and finally, capturing representative mobile AR datasets. 
We hope this paper can shed some lights on how to achieve accurate metric depth estimation for a wide array of mobile AR applications. 

\balance
\scriptsize{
\bibliographystyle{ACM-Reference-Format}
\bibliography{ref.bib}
}
\end{document}